\documentclass{article} 
\usepackage[preprint]{colm2025_conference}
\usepackage{tcolorbox}
\usepackage{microtype}
\usepackage{hyperref}
\usepackage{url}
\usepackage{booktabs}
\usepackage{amssymb}
\usepackage{amsmath}
\usepackage{multirow}
\usepackage{subcaption}
\usepackage{svg}
\usepackage{tikz}
\usepackage{listings}
\usetikzlibrary{arrows.meta, positioning, shapes.geometric}

\definecolor{darkblue}{rgb}{0, 0, 0.5}
\hypersetup{colorlinks=true, citecolor=darkblue, linkcolor=darkblue, urlcolor=darkblue}

\title{From Building Blocks to Planning: Multi-Step Spatial \\ Reasoning in LLMs with Reinforcement Learning}

\author{Amir Tahmasbi, Sadegh Majidi, Kazem Taram \& Aniket Bera \\
Department of Computer Science\\
Purdue University\\
\texttt{\{atahmasb,mmajidiy,kazem,aniketbera\}@purdue.edu}
}

\begin{document}

\ifcolmsubmission
\fi

\maketitle

\begin{abstract}
Spatial reasoning in large language models (LLMs) has gained increasing attention due to applications in navigation and planning. Despite strong general language capabilities, LLMs still struggle with spatial transformations and multi-step planning in structured environments.
We propose a two-stage approach that decomposes spatial reasoning into atomic building blocks and their composition. First, we apply supervised fine-tuning on elementary spatial transformations, such as rotation, translation, and scaling, to equip the model with basic spatial physics. We then freeze this physics-aware model and train lightweight LoRA adapters within the GRPO framework to learn policies that compose these building blocks for multi-step planning in puzzle-based environments, in a closed-loop manner.
To support this pipeline, we synthesize an ASCII-art dataset and construct a corresponding ASCII-based reinforcement learning environment. Our method consistently outperforms baselines, including the generic backbone, physics-aware model, and end-to-end RL models, under both Dynamic environments with explicit state updates and Static environments where the model must rely on its internal state across steps. In addition, the proposed approach converges faster and exhibits more stable training compared to end-to-end reinforcement learning from scratch. Finally, we analyze attention patterns to assess whether fine-tuning induces meaningful improvements in spatial understanding. 
\end{abstract}

\section{Introduction}
Spatial reasoning and understanding represent the capability to reason about spatial relationships of objects and transformations in an environment. This includes understanding relative positions, orientations, distances, and the effects of actions that modify spatial configurations. With the recent advancements of LLMs \citep{minaee2025largelanguagemodelssurvey} and VLMs \citep{li2025surveystateartlarge} in a wide variety of tasks, such as math reasoning \citep{ahn2024largelanguagemodelsmathematical} and vision–language understanding \citep{vis-lan-wu-24}, their spatial reasoning capabilities have gained more attention, with applications such as robotics \citep{kong2025autospatialvisuallanguagereasoningsocial} and language navigation tasks \citep{zhang2024visionandlanguagenavigationtodaytomorrow}. However, their abilities in spatial reasoning have not yet been widely explored \citep{wu2024mindseyellmsvisualizationofthought}. Spatial reasoning tasks can often be mapped into several domains, some of which focus on linguistic and natural language scenarios \citep{stepgame-li-24}, as well as puzzle-based settings \citep{puzzle-lang-noever-21} such as mazes \citep{einarsson2025mazeevalbenchmarktestingsequential}, Rubik's Cube \citep{Ding2023EverythingOT}, and Sokoban \citep{Todd_2023}, where the model needs to understand the spatial relationships among objects to solve the task. In puzzle-based settings, language models are empowered by several techniques. One category involves using an external module as a solver, with the LLM acting as an action recommender or reasoning candidate generator. The external module can be heuristic-based, such as BFS or DFS in Tree of Thoughts (ToT) \citep{yao2023tot}, or learning-based approaches, such as XoT, which leverages pretrained reinforcement learning and Monte Carlo Tree Search (MCTS) \citep{Ding2023EverythingOT}, or Q-learning \citep{deng2025llm_path_planning}. Another category of approaches focuses on the model itself. One direction aims to change the model's behavior through prompting, such as Visualization-of-Thought (VoT) \citep{wu2024mindseyellmsvisualizationofthought}, which elicits spatial reasoning in LLMs by visualizing their reasoning traces and guiding subsequent reasoning steps. Another approach, presented in \citet{dao2025alphamazeenhancinglargelanguage}, is inspired by DeepSeek-R1 \citep{deepseekai2025deepseekr1incentivizingreasoningcapability} and is based on supervised fine-tuning the model on solution traces, and then applying GRPO to further refine reasoning steps on the same task and structure. 

\begin{figure}[t]
\centering
\includegraphics[width=\linewidth]{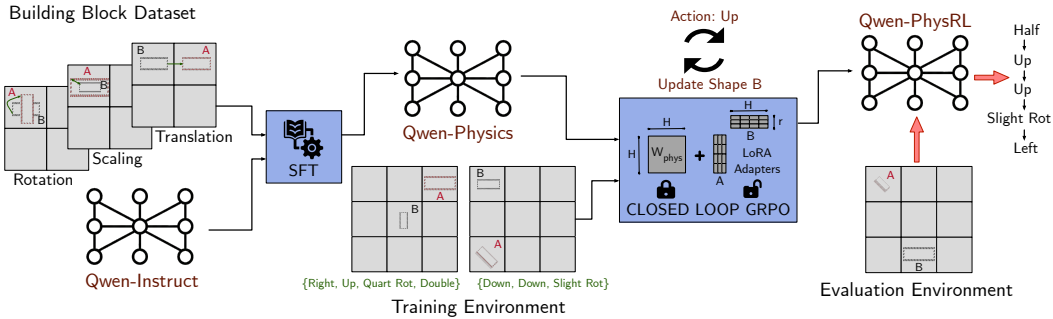}
\caption{\textbf{Overview of the proposed training framework.} The pipeline follows a two-stage learning approach. In the SFT Phase, the base model (\textsc{Qwen-Instruct}) is fine-tuned on the Building Block Dataset to acquire atomic spatial priors (translation, scaling, rotation), resulting in the intermediate \textsc{Qwen-Physics} model. In the RL Phase, we employ GRPO in a closed-loop setting with LoRA adapters. The model is trained to master multi-step spatial reasoning and planning, yielding the final \textsc{Qwen-PhysRL} model.}
\label{fig:pipeline-overview}
\end{figure}

In this work, we focus on spatial reasoning in a puzzle-based setting, where an agent must transform an initial spatial configuration into a target configuration through a sequence of discrete actions. We propose a novel approach that decomposes spatial understanding into a set of building blocks, consisting of atomic transformations such as rotating a shape by $90^\circ$ or translating it one grid cell upward. We first apply supervised fine-tuning to enable the model to learn these basic physical transformations. After this stage, the physics-aware model is kept frozen, and reinforcement learning is applied on top of it by introducing lightweight adapter layers that learn a policy for composing these building blocks as primitives to reach a target spatial configuration from a given starting point. An overview of our approach is shown in Figure~\ref{fig:pipeline-overview}.

For the supervised fine-tuning stage, we synthesize a dataset of 12k tasks spanning three transformation categories, translation, rotation, and scaling, which is used to fine-tune the Qwen2.5-1.5B-Instruct model \citep{qwen2.5-1.5b-instruct}. In the subsequent reinforcement learning stage, the physics-aware model is embedded directly in the reinforcement learning loop within a multi-step, compositional environment. Reinforcement learning is applied via GRPO to optimize lightweight LoRA adapter layers~\citep{hu2021loralowrankadaptationlarge} on top of the frozen backbone, enabling the model to learn policies over sequences of atomic spatial operations through repeated interaction with the environment. We evaluate our approach against several baselines, including the generic Qwen2.5-1.5B-Instruct model, the physics-aware model trained only with supervised fine-tuning, and a Qwen model trained directly with GRPO reinforcement learning. All models are tested on unseen spatial reasoning tasks under two settings: one where the environment map is updated after each action, and one where the map remains fixed. Our results show that the proposed method achieves higher rewards than all baselines across both settings and converges faster during reinforcement learning. Although the physics-aware model trained only with supervised fine-tuning underperforms the generic backbone on the final task, it provides a stronger prior for reinforcement learning, leading to substantially better performance after GRPO optimization. Finally, we conduct ablation studies on attention layers to analyze whether the learned improvements reflect genuine spatial understanding.

\section{Related Work and Background} 
\subsection{Spatial Reasoning in LLMs} 
Spatial reasoning in large language models has been studied across a range of domains. One line of research focuses on textual inputs, where spatial semantics are embedded in text and models are required to reason about relative spatial relations, such as \textit{left of} or \textit{right of}, expressed through language \citep{SpatialRole}. A further step is explored in natural language navigation tasks, where, given a sequence of textual instructions, the model must maintain an implicit spatial state and track its position over multiple steps. In these settings, spatial understanding is inferred from sequential language instructions rather than from explicit spatial representations such as symbolic layouts \citep{wu2024mindseyellmsvisualizationofthought}. Recent efforts explore symbolic environments, particularly ASCII-art representations, which bridge the gap between text and image modalities in applications such as level generation \citep{Todd_2023}. Results show that large language models are capable of recognizing concepts depicted in ASCII art given textual inputs. However, they still exhibit notable limitations and remain far behind human performance in transformation tasks such as translation, rotation, and robustness to noise \citep{bayani2024testingdepthchatgptscomprehension}, as well as in shape recognition \citep{wang2024bothumandetectingchatgpt, jia2025visual}. Prior work also indicates that supervised fine-tuning can improve model accuracy in these settings \citep{jia2025visual}.
\subsection{Attention Mechanisms in Transformer-Based LLMs} 
Since the introduction of the Transformer architecture~\citep{att_all_you_nead}, it has been widely adopted across a broad range of applications. Most notably, it serves as the dominant foundation for large language models (LLMs), enabling efficient natural language processing (NLP) tasks such as understanding and generating long sequences of text~\citep{Devlin2019BERTPO, fewshotBrown}. The key innovation underlying these advances is the self-attention mechanism~\citep{vatMultiModal2021}, which models long-range dependencies among elements in sequential data and enables effective extraction, processing, and generation of structured outputs. Most contemporary LLMs adopt a decoder-only Transformer architecture, which is more suitable for autoregressive text generation. These models process a sequence of tokens through multiple architecturally similar decoder layers, followed by a final linear projection that maps hidden states to the vocabulary space. Each decoder layer consists of several core components, including self-attention, a feed-forward multilayer perceptron (MLP), and residual connections with normalization. With the exception of a small number of element-wise operations (e.g., softmax and dropout), matrix multiplication dominates the computational workload of these components.


During inference, the token sequence is represented as a matrix of hidden state vectors with dimensionality $H$, corresponding to the model's hidden size. These hidden states are multiplied by a set of learned linear projections in the self-attention module, namely $W_Q$ (query), $W_K$ (key), $W_V$ (value), and $W_O$ (output), as well as by projection matrices in the MLP component, such as $W_{\text{up}}$, $W_{\text{down}}$, and $W_{\text{gate}}$. Among these operations, the multiplication $QK^{\mathsf{T}}$, which produces the attention score matrix, plays a key role in our analysis. Here, $Q$ and $K$ are obtained by projecting the input hidden states using $W_Q$ and $W_K$, respectively. The resulting attention score matrix captures semantic and syntactic dependencies between tokens in the sequence by quantifying the relative influence of each token on every other token. In later sections, we examine how this attention distribution changes under our proposed method and analyze its effect on how tokens are weighted during spatial reasoning tasks.

\section{Methodology}
\subsection{Problem Formulation}
We frame the spatial configuration space using three properties: the rotational state of the shape, its translational position on a discrete grid, and its scale. Based on this formulation, a shape state is represented as a tuple \(S = (r, p, s)\), where \(r\) denotes the orientation, \(p\) the spatial position, and \(s\) the size of the shape. Correspondingly, the action space is defined as a finite symbolic set \(\mathcal{A} = \mathcal{A}_{\text{rot}} \cup \mathcal{A}_{\text{trans}} \cup \mathcal{A}_{\text{scale}}\), aligned with these three properties. The rotation action set is given by \(\mathcal{A}_{\text{rot}} = \{90^\circ\ \text{CCW}, 45^\circ\ \text{CCW}, 180^\circ\ \text{CCW}, 0^\circ\}\), the scaling set by \(\mathcal{A}_{\text{scale}} = \{2\times, \tfrac{1}{2}\times, 1\}\), and the translation set by \(\mathcal{A}_{\text{trans}} = \{\text{right}, \text{left}, \text{up}, \text{down}\}\). This discrete action formulation mitigates known limitations of language models when operating in continuous spaces \citep{szot2024groundingmultimodallargelanguage}, while still providing sufficient flexibility to modify each component of the state. The task begins from an initial state \(S_0 = (r_0, p_0, s_0)\), and the agent applies a sequence of actions \(a_{1:T} = \{a_1, \ldots, a_T\}\) with the objective of reaching a target configuration \(S_{\text{target}} = (r_g, p_g, s_g)\). The environment dynamics are deterministic, with state transitions defined as \(S_B^{(t+1)} = \mathcal{T}(S_B^{(t)}, a_t)\), resulting in a closed-loop rollout where the observation at step \(t\) is given by \(O_t = (S_{\text{target}}, S_t, H_t)\), with the action history \(H_t = (a_1, \ldots, a_{t-1})\).\\
The objective is to reach the target spatial configuration in the fewest possible steps. An episode is considered successful once the intersection-over-union (IoU) between the current shape and the target shape exceeds a predefined threshold \(\tau\). Accordingly, the task objective can be expressed in terms of the minimal timestep \(t^\star\) at which the following condition is satisfied:
\begin{equation}
t^\star = \min \left\{\, t \ge 0 \;\middle|\; \operatorname{IoU}\big(S_B^{(t)}, S_A\big) \ge \tau \,\right\}.
\end{equation}
All actions are admissible at each timestep under a bounded map domain \(\mathcal{P} \subset \mathbb{Z}^2\). For any action \(a \in \mathcal{A}\) with induced displacement \(\Delta(a)\), the next position is given by \(\tilde{p} = p + \Delta(a)\) when \(\tilde{p} \in \mathcal{P}\), and equals \(p\) otherwise.

\subsection{Proposed Method}
In the first stage, we perform supervised fine-tuning on atomic building-block transformations. We define a building block as a single-step transformation for which the distance between the start configuration \(S_{\text{start}}\) and the target configuration \(S_{\text{target}}\) satisfies \(\operatorname{dist}(S_{\text{start}}, S_{\text{target}}) = 1\), where the distance is defined as the sum of differences in position, scale, and orientation, i.e., \(\Delta p + \Delta s + \Delta r\). Under this formulation, each training example corresponds to exactly one atomic action from the predefined action set \(\mathcal{A}\). 
This process yields a physics-aware policy \(\pi_{\text{phys}}\), defined as
\begin{equation}
    \pi_{\text{phys}} := \pi_{\theta^*}(a \mid S_{\text{start}}, S_{\text{target}}, k),
\end{equation}
where \(k \in \{\text{rot}, \text{trans}, \text{scale}\}\) denotes the transformation type label. This policy captures the local physics of the environment by learning to correctly execute the corresponding atomic transformation.\\
In the second stage, we learn a compositional policy on top of the frozen physics-aware model. The policy \(\pi_{\phi}\) is parameterized by the frozen base weights \(\theta^*\) obtained from SFT and a set of learnable Low-Rank Adaptation (LoRA) parameters \(\phi\). We apply LoRA to a predefined set of transformer modules \(\mathcal{M} = \{W_Q, W_K, W_V, W_O, W_{\text{gate}}, W_{\text{up}}, W_{\text{down}}\}\). For each layer \(l \in \{1, \ldots, L\}\) and each module \(W \in \mathcal{M}\), the forward pass is modified as
\begin{equation}
    h^{(l)} = (W_{\text{phys}} + \Delta W_l)\, h^{(l-1)} = (W_{\text{phys}} + B_l A_l)\, h^{(l-1)},
\end{equation}
where \(W_{\text{phys}}\) denotes the frozen base weights and \(A_l, B_l\) are the learnable low-rank matrices comprising \(\phi\). The adapter parameters \(\phi\) are optimized using GRPO in a closed-loop reinforcement learning setting, while the base parameters \(\theta^*\) remain fixed.

As a result, the learned policy \(\pi_{\phi}\) operates in a closed-loop manner, where at each timestep \(t\) the model generates a textual output \(y_t\) conditioned on the current observation \(o_t\). 
\begin{equation}
    y_t \sim \pi_{\phi}(\cdot \mid o_t),
    \qquad
    a_t = g(y_t) \in \mathcal{A},
\end{equation}
where \(g(\cdot)\) is a deterministic parser that maps the generated text to a discrete atomic action label.

\section{Experiments}
\subsection{Experimental Setup}
For the spatial reasoning task, all models are built on top of the Qwen2.5-1.5B-Instruct backbone. In our experiments, we evaluate the performance of several variants to isolate the contribution of each training stage. These include: (i) \textbf{ \textsc{Qwen-Instruct}}, the generic pretrained model without task-specific training; (ii) \textbf{\textsc{Qwen-Physics}}, a supervised fine-tuned model trained only on atomic building-block transformations; (iii) \textbf{\textsc{Qwen-DirectRL}}, a model trained end-to-end using GRPO directly on the base model; (iv) \textbf{\textsc{Qwen-PhysRL}}, our proposed two-stage method combining frozen atomic execution with reinforcement learning--based composition; and (v) \textbf{\textsc{Random Policy ($\pi_{rnd}$)}}, a random action policy serving as an unbiased lower-bound baseline. All model variants share the same tokenizer, namely the standard tokenizer provided with the Qwen2.5-1.5B-Instruct model.

\subsection{Dataset Construction}
For supervised fine-tuning, we construct a synthetic dataset consisting of 12k unique samples, approximately uniformly distributed across the action set \(\mathcal{A}\). Each sample is generated by randomly initializing two of the three shape properties and modifying the remaining property according to a single atomic action. All samples are generated programmatically by applying deterministic atomic transformations to randomly sampled initial configurations, ensuring full reproducibility and eliminating the need for manual annotation.
All spatial configurations are represented using an ASCII-art domain, where both the current shape and the target shape are encoded as text-based grids. In this representation, the relative spatial relationship between the current and target shapes is conveyed implicitly through their layouts (Appendix~\S\ref{app:sample-env}). Shape boundaries are denoted using the character \texttt{\#}, while rows are separated using \texttt{*} and newline (\texttt{\textbackslash n}) characters to preserve the two-dimensional structure. This textual spatial encoding follows conventions inspired by prior work on symbolic spatial reasoning with language models~\citep{Todd_2023}. 

\subsection{Evaluation Metrics}
A deterministic parser \(g(\cdot)\) is used to identify the \texttt{<answer>} tags in the model output and extract the corresponding discrete action.
Our primary evaluation metric is the cumulative episode reward. For an episode of length \(T\), we define
\begin{equation}
R_{\text{total}} = R_{\text{correctness}} - 0.1\,T - \sum_{t=1}^{T} R_{\text{rep}}^{(t)} + R_{\text{success}},
\end{equation}

where \(R_{\text{correctness}}\) reflects agreement between the agent's predicted action and a predefined ground-truth sequence \(\mathrm{GT} = \{a_1, \ldots, a_T\}\) of heuristic greedy atomic actions that locally reduce the distance to the target configuration. A per-step penalty of \(0.1\) encourages shorter solutions, \(R_{\text{rep}}^{t}\) penalizes repetitive behaviors, and \(R_{\text{success}}\) is granted upon task completion. We additionally log step-wise rewards \(\{r_t\}_{t=1}^{T}\) to support fine-grained analysis of policy failures and divergence points.

\section{Results and Analysis}
For GRPO training of the physics-aware fine-tuned model, all experiments were conducted on a single NVIDIA A100 GPU with 80~GB of memory. We use the Adam optimizer with a learning rate of \(1\times10^{-5}\) and apply LoRA adaptation with rank \(r=64\) to the specified transformer modules, while keeping the base model frozen. Training is performed with a batch size of 64 trajectories, with puzzles randomly generated at each environment reset. Action sampling during training follows a temperature scheduling strategy to stabilize learning, where the temperature is linearly annealed from an initial value of 1.4 to a lower bound of 0.7 over training iterations. During evaluation, sampling is disabled, and actions are selected greedily to obtain the best deterministic solution from the model. In both training and evaluation, the maximum distance between the start and target configurations is limited to 5, and the episode horizon is set to \(T_{\max}=5\). A success bonus of \(+2\) is assigned upon task completion, while a per-step penalty of \(-0.1\) encourages shorter solutions. To discourage repetitive behavior, an additional repetition penalty of \(-0.2\) is applied when the same atomic action is selected more than twice within an episode. Finally, the correctness reward is normalized to 1 across all operation types, regardless of the number of ground-truth atomic actions available for a given operation, in order to prevent bias toward specific transformation categories. For instance, in a scenario involving three transformations, one rotation and one scaling, the maximum episode reward is computed as \(3 \times \tfrac{1}{3} + 1 + 1 - 3 \times 0.1 = 2.7\). Adding the success bonus \(R_{\text{success}} = 2\) yields a total reward of 4.7.

\begin{figure}[t]
\centering
\includegraphics[width=0.8\linewidth]{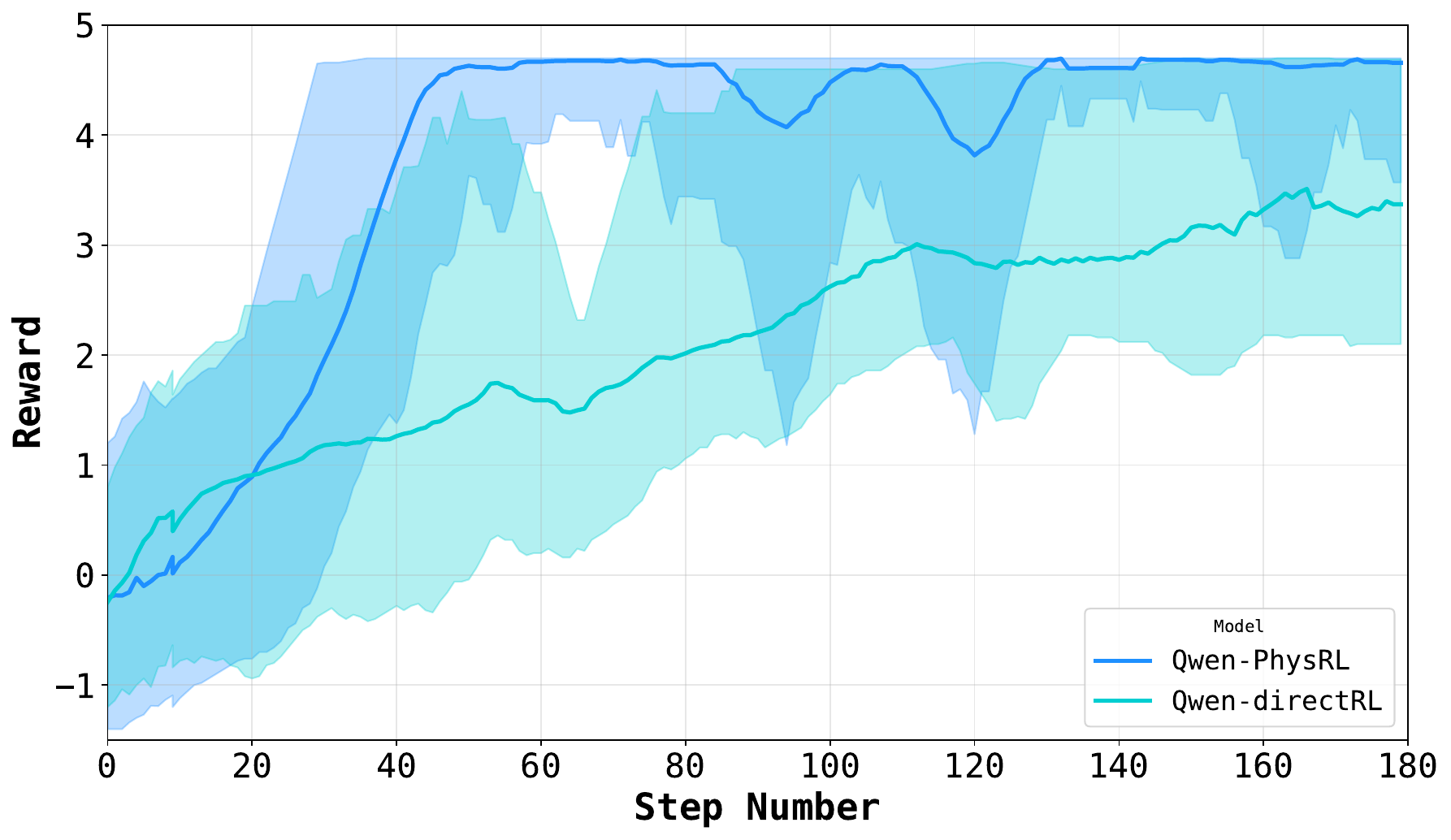}
\caption{GRPO training reward trajectories for \textsc{Qwen-PhysRL} and \textsc{Qwen-DirectRL}, illustrating improved stability and faster convergence when using a frozen physics-aware prior.}
\label{fig:GRPO}
\end{figure}


\begin{table}[t]
\centering
\renewcommand{\arraystretch}{1.3}  
\begin{tabular}{|l|c|c|}
\hline
\multirow{2}{*}{\textbf{Model}} & \multicolumn{2}{c|}{\textbf{Avg $R_{\text{total}}-R_{\text{success}}$  (max = 2.7)}} \\
\cline{2-3}
 & \textbf{Dynamic} & \textbf{Static} \\
\hline
Qwen-Instruct & $0.070$ & $0.004$ \\
Qwen-Physics & $-0.068$ & $-0.120$ \\
Qwen-DirectRL ($r=64$) & $1.626$ & $-0.216$ \\
Qwen-PhysRL ($r=64$, Ours) & $\mathbf{2.457}$ & $\mathbf{1.717}$ \\
\hline
\end{tabular}
\caption{ Performance comparison across models on 100 unseen random scenarios under Dynamic state updates and Static settings.}

\label{tab:rewards}
\end{table}

\paragraph{Performance Analysis.}
The average total reward achieved by different model variants is reported in Table~\ref{tab:rewards}. We evaluate all models under two settings. In the \textbf{Dynamic} setting, after each action, the environment updates the shape's spatial configuration map $S_B$, mirroring the training procedure and generating the next prompt accordingly. This setup allows the model to focus on predicting the correct next action without requiring explicit memorization of prior state changes. In contrast, in the \textbf{Static} setting, the initial maps remain fixed, and the prompt includes only the sequence of previously selected actions. Successful performance in this scenario, therefore, requires the model to internally track and reason about the cumulative effects of past actions. The results demonstrate that our proposed \textsc{Qwen-PhysRL} achieves near-maximum performance in the Dynamic setting, with an average reward of $2.457$, indicating that the learned RL policy reliably identifies the correct sequence of transformations when external state updates are provided. Importantly, \textsc{Qwen-PhysRL} also achieves an average reward of $1.717$ in the Static setting, substantially outperforming all baseline models. In contrast, \textsc{Qwen-DirectRL} exhibits moderate performance in the Dynamic setting but fails in the Static setting, suggesting that reinforcement learning alone is insufficient to equip the model with robust internal spatial understanding and reasoning. 
Finally, non-RL models perform poorly in both settings, reflecting limited planning capability. Reinforcement learning bridges this gap by teaching the model how to utilize its building-block knowledge, enabling the composition of atomic operations into coherent multi-step policies.

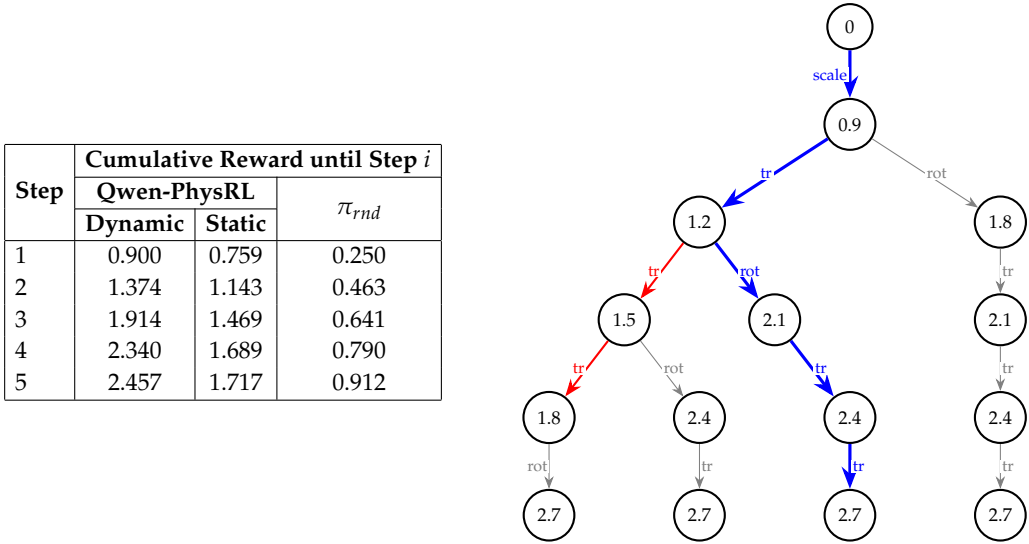
\begin{figure}[t]
\centering
\begin{minipage}{0.45\textwidth}
\centering
\footnotesize
\renewcommand{\arraystretch}{1.2}
\setlength{\tabcolsep}{4pt} 
\begin{tabular}{|l|c|c|c|}
\hline
\multirow{3}{*}{\textbf{Step}} & \multicolumn{3}{c|}{\textbf{Cumulative Reward until Step $i$}} \\
\cline{2-4}
 & \multicolumn{2}{c|}{\textbf{Qwen-PhysRL}} & \multirow{2}{*}{\textbf{$\pi_{rnd}$}} \\
\cline{2-3}
 & \textbf{Dynamic} & \textbf{Static} & \\
\hline
1 & 0.900 & 0.759 & 0.250 \\
2 & 1.374 & 1.143 & 0.463 \\
3 & 1.914 & 1.469 & 0.641 \\
4 & 2.340 & 1.689 & 0.790 \\
5 & 2.457 & 1.717 & 0.912 \\
\hline
\end{tabular}
\end{minipage}%
\hfill
\begin{minipage}{0.5\textwidth}
\centering
\begin{tikzpicture}[
    node distance=0.7cm,
    every node/.style={font=\scriptsize},
    state/.style={circle, draw, thick, minimum size=0.6cm},
    >=Stealth,
    edge_label/.style={fill=white, inner sep=0.5pt, font=\tiny}
]
\def\levelspace{1.3cm}
\def\hspacing{2}  

\node[state] (s0) at (0,0) {0};

\node[state] (s1a) at (0,-\levelspace) {0.9};

\node[state] (s2a) at (-1*\hspacing,-2*\levelspace) {1.2};
\node[state] (s2b) at (1*\hspacing,-2*\levelspace) {1.8};

\node[state] (s3a) at (-1.5*\hspacing,-3*\levelspace) {1.5};
\node[state] (s3b) at (-0.5*\hspacing,-3*\levelspace) {2.1};
\node[state] (s3c) at (1*\hspacing,-3*\levelspace) {2.1};

\node[state] (s4a) at (-2*\hspacing,-4*\levelspace) {1.8};
\node[state] (s4b) at (-1*\hspacing,-4*\levelspace) {2.4};
\node[state] (s4c) at (0,-4*\levelspace) {2.4};
\node[state] (s4d) at (1*\hspacing,-4*\levelspace) {2.4};

\node[state] (s5a) at (-2*\hspacing,-5*\levelspace) {2.7};
\node[state] (s5b) at (-1*\hspacing,-5*\levelspace) {2.7};
\node[state] (s5c) at (0,-5*\levelspace) {2.7};
\node[state] (s5d) at (1*\hspacing,-5*\levelspace) {2.7};

\draw[->, very thick, blue] (s0) -- node[edge_label, left] {scale} (s1a);
\draw[->, very thick, blue] (s1a) -- node[edge_label, left] {tr} (s2a);
\draw[->, very thick, blue] (s2a) -- node[edge_label, right] {rot} (s3b);
\draw[->, very thick, blue] (s3b) -- node[edge_label, right] {tr} (s4c);
\draw[->, very thick, blue] (s4c) -- node[edge_label, right] {tr} (s5c);

\draw[->, gray] (s1a) -- node[edge_label, right] {rot} (s2b);
\draw[->, gray] (s2b) -- node[edge_label, right] {tr} (s3c);
\draw[->, gray] (s3c) -- node[edge_label, right] {tr} (s4d);
\draw[->, gray] (s4d) -- node[edge_label, right] {tr} (s5d);
\draw[->, thick, red] (s2a) -- node[edge_label, left] {tr} (s3a);
\draw[->, thick, red] (s3a) -- node[edge_label, left] {tr} (s4a);
\draw[->, gray] (s3a) -- node[edge_label, right] {rot} (s4b);
\draw[->, gray] (s4a) -- node[edge_label, left] {rot} (s5a);
\draw[->, gray] (s4b) -- node[edge_label, right] {tr} (s5b);
\end{tikzpicture}
\end{minipage}

\caption{Left: Step-by-step cumulative reward comparison between \textsc{Qwen-PhysRL} (Dynamic and Static settings) and random policy ($\pi_{rnd}$) on tasks requiring 3 translations, 1 rotation, and 1 scaling. Right: Action-reward trajectory illustrating the optimal path (blue: \texttt{scale} $\rightarrow$ \texttt{tr} $\rightarrow$ \texttt{rot} $\rightarrow$ \texttt{tr} $\rightarrow$ \texttt{tr}), which achieves the maximum reward of 2.7, contrasted with alternative action sequences (gray). The trajectory obtained under the \emph{Static} setting is shown in red, where the model initially follows the optimal prefix (\texttt{scale} $\rightarrow$ \texttt{tr} $\rightarrow$ \texttt{tr} $\rightarrow$ \texttt{tr}) but then diverges and loses track of the remaining steps, failing to complete the full plan.
}

\label{fig:reward_structure}
\end{figure}

\paragraph{Per-Step Reward Analysis.}
To analyze step-by-step reward accumulation during evaluation, we focus on a smaller subset of test samples in which the distance between the start and target configurations is exactly five. Specifically, we select instances that require three translations, one rotation, and one scaling operation to reach the goal. This restriction allows us to study reward trajectories under a fixed action budget and comparable difficulty. Figure~\ref{fig:reward_structure} reports the average cumulative reward per step for three cases: \textsc{Qwen-PhysRL} evaluated in the Dynamic environment, \textsc{Qwen-PhysRL} evaluated in the Static setting, and a random policy $\pi_{\text{rnd}}$. Figure~\ref{fig:reward_structure} also includes a tree representation of the maximum achievable cumulative reward at each step for all valid sequences of ground-truth actions ($GT$). To keep the tree visualization compact and interpretable, we fix the first action in the tree to \texttt{scale}, which is the most frequently selected initial operation by the learned policy. The optimal action sequence identified by the \textsc{Qwen-PhysRL} policy in the Dynamic setting (\texttt{scale} $\rightarrow$ \texttt{translation} $\rightarrow$ \texttt{rotation} $\rightarrow$ \texttt{translation} $\rightarrow$ \texttt{translation}) is highlighted as the blue path in the tree and is taken in approximately 33~\% of the evaluated trajectories. The corresponding average cumulative reward closely tracks the maximum achievable reward along this path, indicating that the model consistently follows near-optimal action sequences when the environment explicitly updates the state after each step. Two additional high-probability paths, (\texttt{scale} $\rightarrow$ \texttt{rotation} $\rightarrow$ \texttt{translation} $\rightarrow$ \texttt{translation} $\rightarrow$ \texttt{translation}) and (\texttt{scale} $\rightarrow$ \texttt{translation} $\rightarrow$ \texttt{translation} $\rightarrow$ \texttt{rotation} $\rightarrow$ \texttt{translation}), are taken in 27~\% and 19~\% of cases, respectively. In contrast, under the Static setting, the average cumulative reward initially follows the red path in the tree, corresponding to the sequence (\texttt{scale} $\rightarrow$ \texttt{translation} $\rightarrow$ \texttt{translation} $\rightarrow$ \texttt{translation}). This behavior aligns with our empirical observations that, without explicit state updates, the model gradually loses track of earlier actions, diverges from the optimal plan, and ultimately fails to produce a decisive fifth action. As a result, reward accumulation stagnates at later steps. Compared to the random policy $\pi_{\text{rnd}}$ (see Appendix~\S\ref{app:rand} for its expected reward analysis), \textsc{Qwen-PhysRL} selects substantially more effective actions at each step in both the Dynamic and Static settings. The only exception occurs at the final step in the Static case, where the learned policy yields a smaller average reward increase than the random policy, as it has already diverged from the optimal trajectory, whereas the random policy occasionally converges to a higher-reward branch by chance. We note that, because the action space is not relatively large and the reward distribution is partly biased toward positive values, the random policy attains relatively high rewards in this setting. Even under these favorable conditions, \textsc{Qwen-PhysRL} consistently outperforms the random baseline by a wide margin. 
\begin{figure}[t]
\centering
\includegraphics[width=1\linewidth]{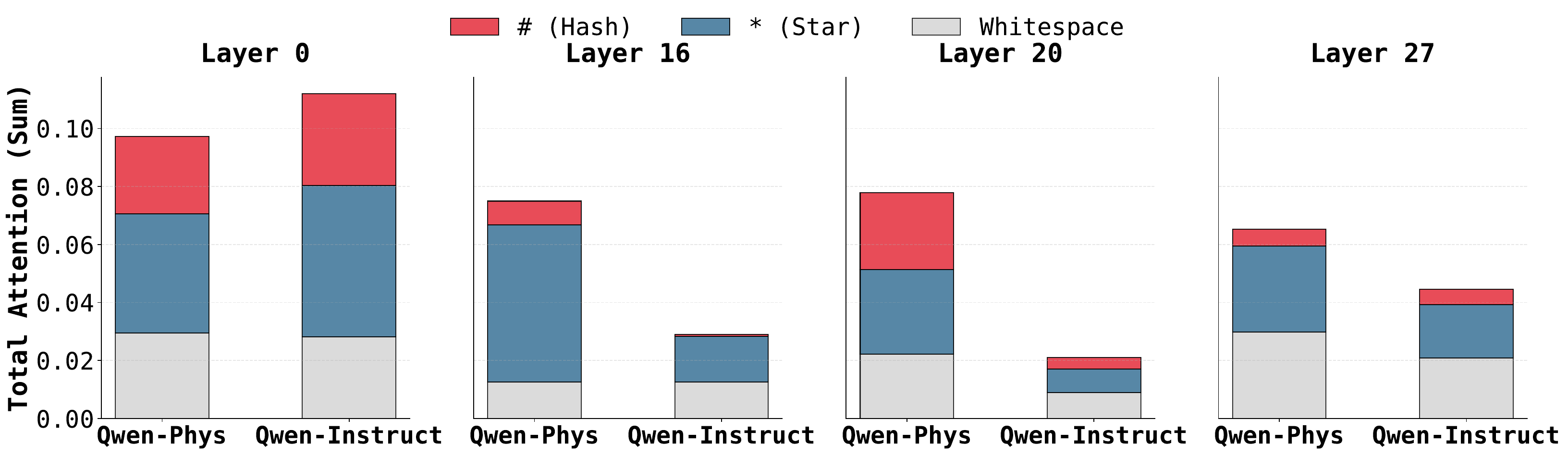}
\caption{Token-level attention distribution across layers for \textsc{Qwen-Physics} and \textsc{Qwen-Instruct}.}
\label{fig:attention}
\end{figure}
\paragraph{GRPO Convergence and Stability.}
As shown in Figure~\ref{fig:GRPO}, \textsc{Qwen-PhysRL} exhibits substantially faster convergence and higher final rewards than \textsc{Qwen-DirectRL} under identical GRPO training configurations, highlighting the importance of a physics-aware initialization for RL. Table~\ref{tab:rewards} provides additional insight: the generic \textsc{Qwen-Instruct} backbone used in \textsc{Qwen-DirectRL} achieves higher rewards than the \textsc{Qwen-Physics} backbone used in \textsc{Qwen-PhysRL} during evaluation, indicating that general pretrained knowledge can partially compensate. However, despite this advantage, \textsc{Qwen-Instruct} fails to scale effectively under GRPO, exhibiting slower adaptation and early convergence to suboptimal policies. In contrast, the proposed two-stage approach consistently achieves superior performance across both Dynamic and Static settings.

\subsection{Ablation Study}
\begin{figure}[t]
\centering
\begin{subfigure}[b]{0.47\linewidth}
    \centering
    \includegraphics[width=0.9\linewidth]{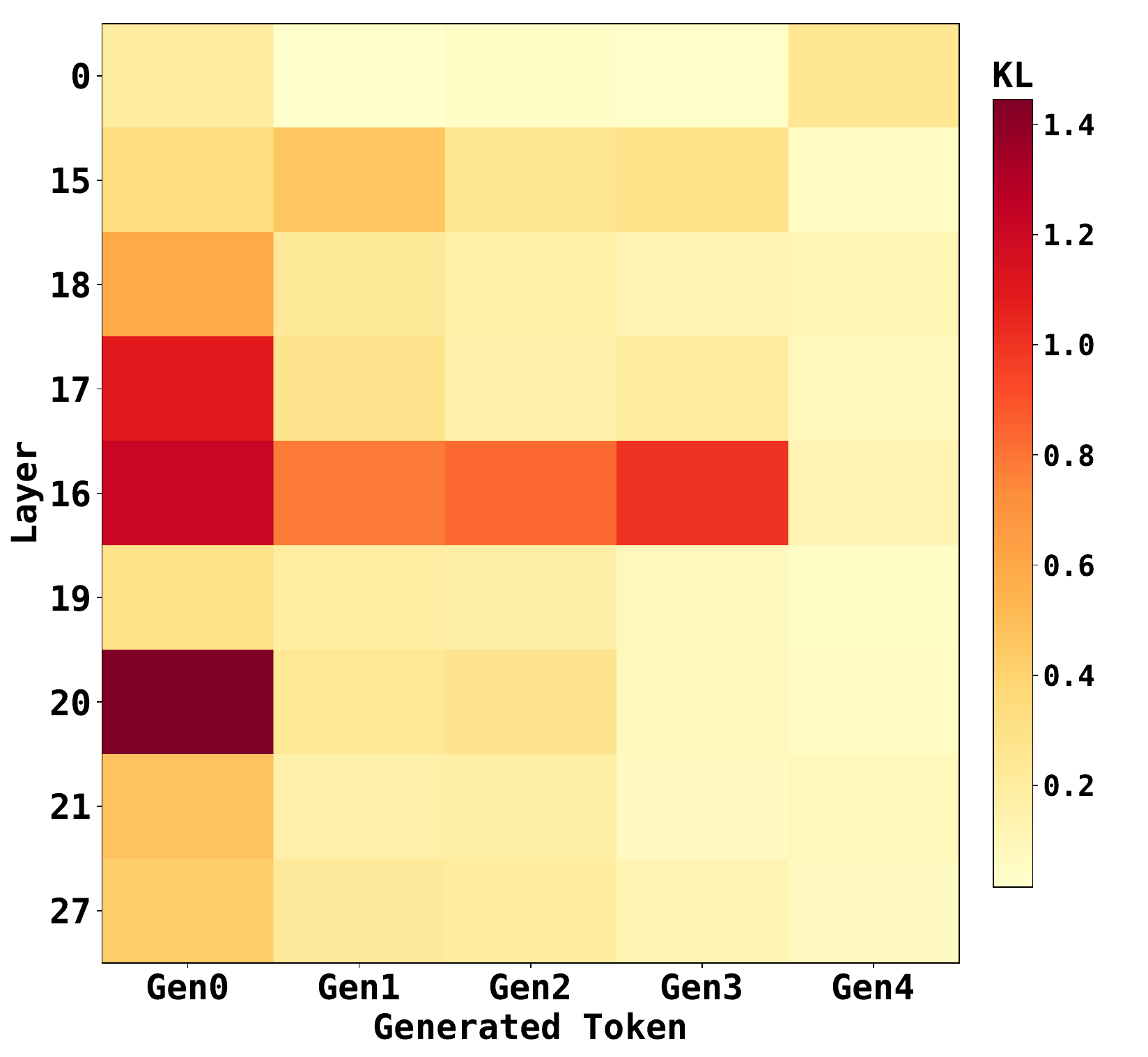}
    \caption{Attention divergence heatmap}
    \label{fig:kl}
\end{subfigure}
\hfill
\begin{subfigure}[b]{0.52\linewidth}
    \centering
    \includegraphics[width=1\linewidth]{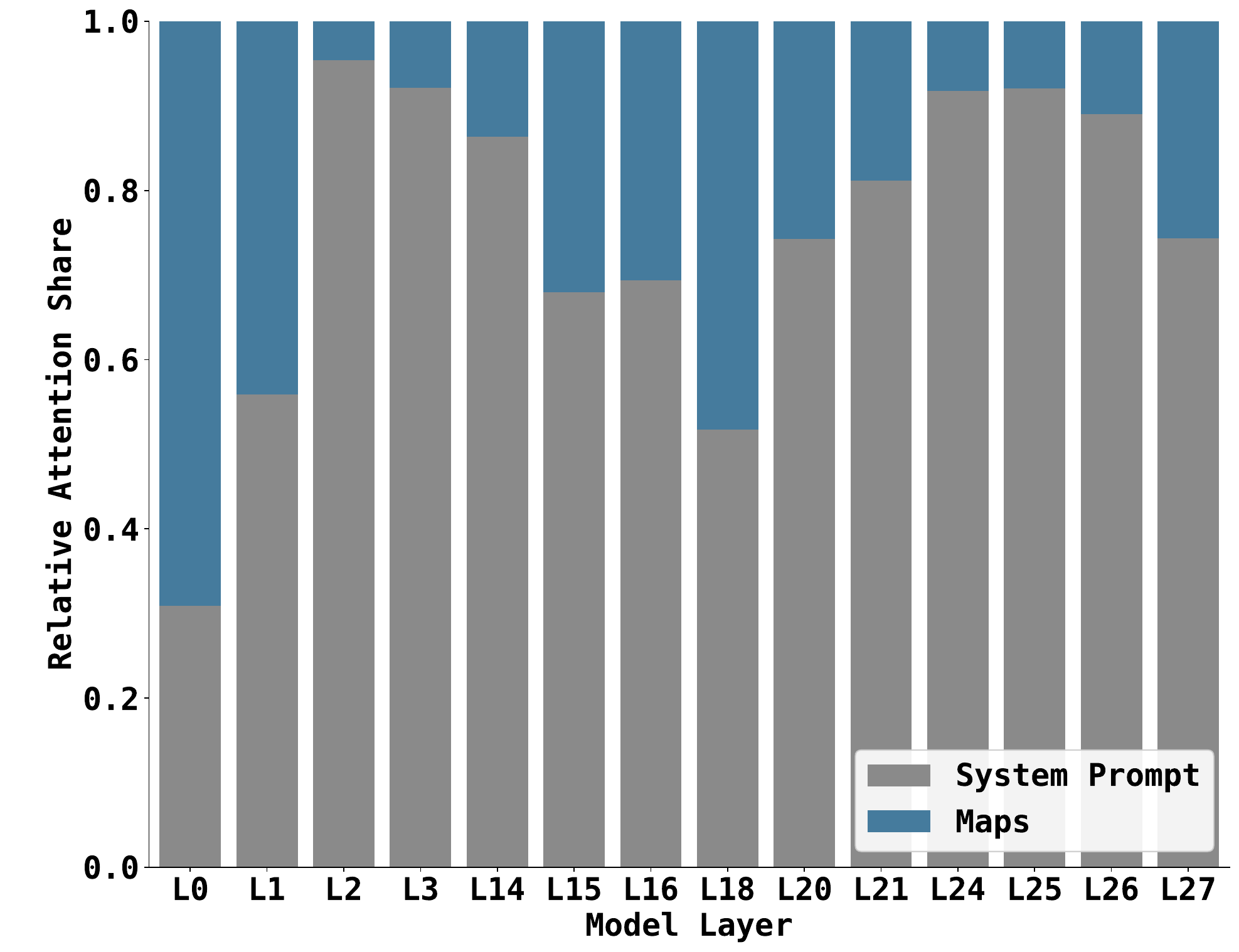}
    \caption{Relative attention share across layers}
    \label{fig:prompt}
\end{subfigure}
\caption{Layer-wise attention score analysis showing (a) attention divergence of \textsc{Qwen-Physics} from the backbone \textsc{Qwen-Instruct} model across layers, and (b) relative attention allocation between system prompt and map tokens.}

\label{fig:KL-prompt}
\end{figure}
To examine the effect of our initial building-block fine-tuning on the spatial understanding of the target LLM $\mathcal{F}$, we analyze how attention is distributed across prompt tokens at different layers of the model by studying the attention score matrices produced during evaluation. We prompt the model with a fixed system prompt $P$, concatenated with a start--target map sequence $M$, to generate an answer sequence $A$, i.e., $\mathcal{F}(P \!\cdot\! M) = A$, where $\cdot$ denotes string concatenation. To obtain attention scores between prompt tokens and the newly generated output tokens, we concatenate the generated answer to the original prompt and perform a single forward pass over the full sequence, $\mathcal{F}(P \!\cdot\! M \!\cdot\! A)$. This pass yields the complete set of attention score matrices, including those corresponding to output tokens. For each decoder layer $l \in \{1,\dots,L\}$ and attention head $h \in \{1,\dots,H\}$, the model computes a raw attention score matrix as $Q_{h,l} K_{h,l}^{\mathsf{T}}$. We aggregate these matrices across heads by averaging to obtain a per-layer attention score matrix,
\[
\text{AttScore}_l = \frac{1}{H} \sum_{h=1}^{H} Q_{h,l} K_{h,l}^{\mathsf{T}}.
\]
We then focus on the rows corresponding to the answer tokens, $\text{AttScore}_l[|P \!\cdot\! M|:, :]$, where each row represents the attention assigned to all prompt tokens during the generation of a specific output token. In the following, we summarize key observations derived from analyzing these attention patterns across layers.

\paragraph{Observation 1.}
Fine-tuning induces the most pronounced changes in attention distributions within the middle layers of the model, as confirmed by Figure~\ref{fig:kl}. Among the 28 layers, layers 16 and 20 exhibit the highest KL divergence between \textsc{Qwen-Physics} and the generic \textsc{Qwen-Instruct} backbone, while the early (layer 0) and final (layer 27) layers show substantially smaller deviations. This indicates that the adaptation introduced by fine-tuning is concentrated in intermediate layers rather than being uniformly distributed across the network.

\paragraph{Observation 2.}
\textsc{Qwen-Physics} assigns closer attention to spatially informative ASCII tokens, such as \texttt{\#} and \texttt{*}, compared to \textsc{Qwen-Instruct}. As shown in Figure~\ref{fig:attention}, the increased attention to these tokens is most prominent in the same middle layers identified in Observation~1. This suggests that the layers undergoing the largest adaptation are also those most responsible for encoding spatial structure, enabling the model to better focus on salient regions of the map $M$ during spatial reasoning.

\paragraph{Observation 3.}
Across most layers, \textsc{Qwen-Physics} consistently allocates more attention to the system prompt \(P\) than to the map region, except in the earliest layers. Notably, the system prompt contains approximately \(1.8\times\) more tokens than the map region $M$, yet Figure~\ref{fig:prompt} shows an attention gap that is substantially larger than what token count alone would suggest. This gap narrows in the middle layers, which, as shown in Observations~1 and~2, are more actively involved in spatial reasoning. This behavior is consistent with prior findings that textual context can dominate attention allocation in spatial reasoning and vision--language models~\citep{chen2025spatialreasoninghardvlms, wang2025textspeaksloudervision}. The earliest layers, such as layer 0, allocate attention more evenly across different parts of the input, reflecting a coarse, global skimming behavior, as illustrated in both Figure~\ref{fig:prompt} and Figure~\ref{fig:attention}.

\section{Conclusion}

This paper presents a two-stage training pipeline that enables spatial understanding, reasoning, and planning capabilities in large language models. Our approach first builds a foundation of atomic spatial relations through supervised fine-tuning and then augments it with multi-step planning ability using a closed-loop GRPO-based reinforcement learning stage. This combination equips the model with both task-relevant physical knowledge and a learned policy that effectively composes these primitives to solve more complex spatial planning problems, without relying solely on reinforcement learning to discover representations and planning strategies simultaneously. Experimental results demonstrate that models trained under this framework exhibit a newfound understanding of spatial properties and planning, substantially outperforming generic LLM baselines on challenging spatial planning tasks. Moreover, compared to end-to-end reinforcement learning applied to an unadapted backbone model, our pipeline converges faster and exhibits more stable training under identical settings. Through ablation studies and attention analysis, we further identify the decoder layers most affected by training and show that our approach systematically shifts attention toward task-critical tokens, providing evidence of more structured internal reasoning. While our evaluation focuses on a specific class of spatial planning problems, we believe that the underlying principle of learning reusable building blocks and composing them through reinforcement learning is broadly applicable. We view this work as a step toward more modular and interpretable approaches for teaching complex reasoning skills to language models, and we leave the exploration of additional tasks and modalities to future work.

\bibliography{colm2025_conference}

\begin{thebibliography}{29}
\providecommand{\natexlab}[1]{#1}
\providecommand{\url}[1]{\texttt{#1}}
\expandafter\ifx\csname urlstyle\endcsname\relax
  \providecommand{\doi}[1]{doi: #1}\else
  \providecommand{\doi}{doi: \begingroup \urlstyle{rm}\Url}\fi

\bibitem[Ahn et~al.(2024)Ahn, Verma, Lou, Liu, Zhang, and Yin]{ahn2024largelanguagemodelsmathematical}
Janice Ahn, Rishu Verma, Renze Lou, Di~Liu, Rui Zhang, and Wenpeng Yin.
\newblock Large language models for mathematical reasoning: Progresses and challenges, 2024.
\newblock URL \url{https://arxiv.org/abs/2402.00157}.

\bibitem[Akbari et~al.(2021)Akbari, Yuan, Qian, Chuang, Chang, Cui, and Gong]{vatMultiModal2021}
Hassan Akbari, Liangzhe Yuan, Rui Qian, Wei-Hong Chuang, Shih-Fu Chang, Yin Cui, and Boqing Gong.
\newblock Vatt: transformers for multimodal self-supervised learning from raw video, audio and text.
\newblock In \emph{Proceedings of the 35th International Conference on Neural Information Processing Systems}, NIPS '21, Red Hook, NY, USA, 2021. Curran Associates Inc.
\newblock ISBN 9781713845393.

\bibitem[Bayani(2024)]{bayani2024testingdepthchatgptscomprehension}
David Bayani.
\newblock Testing the depth of chatgpt's comprehension via cross-modal tasks based on ascii-art: Gpt3.5's abilities in regard to recognizing and generating ascii-art are not totally lacking, 2024.
\newblock URL \url{https://arxiv.org/abs/2307.16806}.

\bibitem[Brown et~al.(2020)Brown, Mann, Ryder, Subbiah, Kaplan, Dhariwal, Neelakantan, Shyam, Sastry, Askell, Agarwal, Herbert-Voss, Krueger, Henighan, Child, Ramesh, Ziegler, Wu, Winter, Hesse, Chen, Sigler, Litwin, Gray, Chess, Clark, Berner, McCandlish, Radford, Sutskever, and Amodei]{fewshotBrown}
Tom Brown, Benjamin Mann, Nick Ryder, Melanie Subbiah, Jared~D Kaplan, Prafulla Dhariwal, Arvind Neelakantan, Pranav Shyam, Girish Sastry, Amanda Askell, Sandhini Agarwal, Ariel Herbert-Voss, Gretchen Krueger, Tom Henighan, Rewon Child, Aditya Ramesh, Daniel Ziegler, Jeffrey Wu, Clemens Winter, Chris Hesse, Mark Chen, Eric Sigler, Mateusz Litwin, Scott Gray, Benjamin Chess, Jack Clark, Christopher Berner, Sam McCandlish, Alec Radford, Ilya Sutskever, and Dario Amodei.
\newblock Language models are few-shot learners.
\newblock In H.~Larochelle, M.~Ranzato, R.~Hadsell, M.F. Balcan, and H.~Lin (eds.), \emph{Advances in Neural Information Processing Systems}, volume~33, pp.\  1877--1901. Curran Associates, Inc., 2020.
\newblock URL \url{https://proceedings.neurips.cc/paper_files/paper/2020/file/1457c0d6bfcb4967418bfb8ac142f64a-Paper.pdf}.

\bibitem[Chen et~al.(2025)Chen, Zhu, Zhou, Zhang, Gao, Niebles, Geva, He, Wu, and Li]{chen2025spatialreasoninghardvlms}
Shiqi Chen, Tongyao Zhu, Ruochen Zhou, Jinghan Zhang, Siyang Gao, Juan~Carlos Niebles, Mor Geva, Junxian He, Jiajun Wu, and Manling Li.
\newblock Why is spatial reasoning hard for vlms? an attention mechanism perspective on focus areas, 2025.
\newblock URL \url{https://arxiv.org/abs/2503.01773}.

\bibitem[Dao \& Vu(2025)Dao and Vu]{dao2025alphamazeenhancinglargelanguage}
Alan Dao and Dinh~Bach Vu.
\newblock Alphamaze: Enhancing large language models' spatial intelligence via grpo, 2025.
\newblock URL \url{https://arxiv.org/abs/2502.14669}.

\bibitem[DeepSeek-AI et~al.(2025)DeepSeek-AI, Guo, Yang, Zhang, Song, Zhang, Xu, Zhu, Ma, Wang, Bi, Zhang, Yu, Wu, Wu, Gou, Shao, Li, Gao, Liu, Xue, Wang, Wu, Feng, Lu, Zhao, Deng, Zhang, Ruan, Dai, Chen, Ji, Li, Lin, Dai, Luo, Hao, Chen, Li, Zhang, Bao, Xu, Wang, Ding, Xin, Gao, Qu, Li, Guo, Li, Wang, Chen, Yuan, Qiu, Li, Cai, Ni, Liang, Chen, Dong, Hu, Gao, Guan, Huang, Yu, Wang, Zhang, Zhao, Wang, Zhang, Xu, Xia, Zhang, Zhang, Tang, Li, Wang, Li, Tian, Huang, Zhang, Wang, Chen, Du, Ge, Zhang, Pan, Wang, Chen, Jin, Chen, Lu, Zhou, Chen, Ye, Wang, Yu, Zhou, Pan, Li, Zhou, Wu, Ye, Yun, Pei, Sun, Wang, Zeng, Zhao, Liu, Liang, Gao, Yu, Zhang, Xiao, An, Liu, Wang, Chen, Nie, Cheng, Liu, Xie, Liu, Yang, Li, Su, Lin, Li, Jin, Shen, Chen, Sun, Wang, Song, Zhou, Wang, Shan, Li, Wang, Wei, Zhang, Xu, Li, Zhao, Sun, Wang, Yu, Zhang, Shi, Xiong, He, Piao, Wang, Tan, Ma, Liu, Guo, Ou, Wang, Gong, Zou, He, Xiong, Luo, You, Liu, Zhou, Zhu, Xu, Huang, Li, Zheng, Zhu, Ma, Tang, Zha, Yan, Ren, Ren, Sha, Fu, Xu, Xie, Zhang,
  Hao, Ma, Yan, Wu, Gu, Zhu, Liu, Li, Xie, Song, Pan, Huang, Xu, Zhang, and Zhang]{deepseekai2025deepseekr1incentivizingreasoningcapability}
DeepSeek-AI, Daya Guo, Dejian Yang, Haowei Zhang, Junxiao Song, Ruoyu Zhang, Runxin Xu, Qihao Zhu, Shirong Ma, Peiyi Wang, Xiao Bi, Xiaokang Zhang, Xingkai Yu, Yu~Wu, Z.~F. Wu, Zhibin Gou, Zhihong Shao, Zhuoshu Li, Ziyi Gao, Aixin Liu, Bing Xue, Bingxuan Wang, Bochao Wu, Bei Feng, Chengda Lu, Chenggang Zhao, Chengqi Deng, Chenyu Zhang, Chong Ruan, Damai Dai, Deli Chen, Dongjie Ji, Erhang Li, Fangyun Lin, Fucong Dai, Fuli Luo, Guangbo Hao, Guanting Chen, Guowei Li, H.~Zhang, Han Bao, Hanwei Xu, Haocheng Wang, Honghui Ding, Huajian Xin, Huazuo Gao, Hui Qu, Hui Li, Jianzhong Guo, Jiashi Li, Jiawei Wang, Jingchang Chen, Jingyang Yuan, Junjie Qiu, Junlong Li, J.~L. Cai, Jiaqi Ni, Jian Liang, Jin Chen, Kai Dong, Kai Hu, Kaige Gao, Kang Guan, Kexin Huang, Kuai Yu, Lean Wang, Lecong Zhang, Liang Zhao, Litong Wang, Liyue Zhang, Lei Xu, Leyi Xia, Mingchuan Zhang, Minghua Zhang, Minghui Tang, Meng Li, Miaojun Wang, Mingming Li, Ning Tian, Panpan Huang, Peng Zhang, Qiancheng Wang, Qinyu Chen, Qiushi Du, Ruiqi Ge, Ruisong
  Zhang, Ruizhe Pan, Runji Wang, R.~J. Chen, R.~L. Jin, Ruyi Chen, Shanghao Lu, Shangyan Zhou, Shanhuang Chen, Shengfeng Ye, Shiyu Wang, Shuiping Yu, Shunfeng Zhou, Shuting Pan, S.~S. Li, Shuang Zhou, Shaoqing Wu, Shengfeng Ye, Tao Yun, Tian Pei, Tianyu Sun, T.~Wang, Wangding Zeng, Wanjia Zhao, Wen Liu, Wenfeng Liang, Wenjun Gao, Wenqin Yu, Wentao Zhang, W.~L. Xiao, Wei An, Xiaodong Liu, Xiaohan Wang, Xiaokang Chen, Xiaotao Nie, Xin Cheng, Xin Liu, Xin Xie, Xingchao Liu, Xinyu Yang, Xinyuan Li, Xuecheng Su, Xuheng Lin, X.~Q. Li, Xiangyue Jin, Xiaojin Shen, Xiaosha Chen, Xiaowen Sun, Xiaoxiang Wang, Xinnan Song, Xinyi Zhou, Xianzu Wang, Xinxia Shan, Y.~K. Li, Y.~Q. Wang, Y.~X. Wei, Yang Zhang, Yanhong Xu, Yao Li, Yao Zhao, Yaofeng Sun, Yaohui Wang, Yi~Yu, Yichao Zhang, Yifan Shi, Yiliang Xiong, Ying He, Yishi Piao, Yisong Wang, Yixuan Tan, Yiyang Ma, Yiyuan Liu, Yongqiang Guo, Yuan Ou, Yuduan Wang, Yue Gong, Yuheng Zou, Yujia He, Yunfan Xiong, Yuxiang Luo, Yuxiang You, Yuxuan Liu, Yuyang Zhou, Y.~X. Zhu,
  Yanhong Xu, Yanping Huang, Yaohui Li, Yi~Zheng, Yuchen Zhu, Yunxian Ma, Ying Tang, Yukun Zha, Yuting Yan, Z.~Z. Ren, Zehui Ren, Zhangli Sha, Zhe Fu, Zhean Xu, Zhenda Xie, Zhengyan Zhang, Zhewen Hao, Zhicheng Ma, Zhigang Yan, Zhiyu Wu, Zihui Gu, Zijia Zhu, Zijun Liu, Zilin Li, Ziwei Xie, Ziyang Song, Zizheng Pan, Zhen Huang, Zhipeng Xu, Zhongyu Zhang, and Zhen Zhang.
\newblock Deepseek-r1: Incentivizing reasoning capability in llms via reinforcement learning, 2025.
\newblock URL \url{https://arxiv.org/abs/2501.12948}.

\bibitem[Deng et~al.(2025)Deng, Zhang, Ou, and Feng]{deng2025llm_path_planning}
Hourui Deng, Hongjie Zhang, Jie Ou, and Chaosheng Feng.
\newblock Can llm be a good path planner based on prompt engineering? mitigating the hallucination for path planning.
\newblock In \emph{Proceedings of the International Conference on Intelligent Computing (ICIC)}, 2025.

\bibitem[Devlin et~al.(2019)Devlin, Chang, Lee, and Toutanova]{Devlin2019BERTPO}
Jacob Devlin, Ming-Wei Chang, Kenton Lee, and Kristina Toutanova.
\newblock Bert: Pre-training of deep bidirectional transformers for language understanding.
\newblock In \emph{North American Chapter of the Association for Computational Linguistics}, 2019.
\newblock URL \url{https://api.semanticscholar.org/CorpusID:52967399}.

\bibitem[Ding et~al.(2023)Ding, Zhang, Wang, Xu, Ma, Zhang, Qin, Rajmohan, Lin, and Zhang]{Ding2023EverythingOT}
Ruomeng Ding, Chaoyun Zhang, Lu~Wang, Yong Xu, Ming-Jie Ma, Wei Zhang, Si~Qin, S.~Rajmohan, Qingwei Lin, and Dongmei Zhang.
\newblock Everything of thoughts: Defying the law of penrose triangle for thought generation.
\newblock \emph{ArXiv}, abs/2311.04254, 2023.

\bibitem[Einarsson(2025)]{einarsson2025mazeevalbenchmarktestingsequential}
Hafsteinn Einarsson.
\newblock Mazeeval: A benchmark for testing sequential decision-making in language models, 2025.
\newblock URL \url{https://arxiv.org/abs/2507.20395}.

\bibitem[Hu et~al.(2021)Hu, Shen, Wallis, Allen-Zhu, Li, Wang, Wang, and Chen]{hu2021loralowrankadaptationlarge}
Edward~J. Hu, Yelong Shen, Phillip Wallis, Zeyuan Allen-Zhu, Yuanzhi Li, Shean Wang, Lu~Wang, and Weizhu Chen.
\newblock Lora: Low-rank adaptation of large language models, 2021.
\newblock URL \url{https://arxiv.org/abs/2106.09685}.

\bibitem[Jia et~al.(2025)Jia, Yue, Huang, Qin, Liu, Lin, and You]{jia2025visual}
Qi~Jia, Xiang Yue, Shanshan Huang, Ziheng Qin, Yizhu Liu, Bill~Yuchen Lin, and Yang You.
\newblock Visual perception in text strings, 2025.
\newblock URL \url{https://openreview.net/forum?id=etToTig9Fp}.

\bibitem[Kong et~al.(2025)Kong, Song, Liang, Manocha, Yao, and Xiao]{kong2025autospatialvisuallanguagereasoningsocial}
Yangzhe Kong, Daeun Song, Jing Liang, Dinesh Manocha, Ziyu Yao, and Xuesu Xiao.
\newblock Autospatial: Visual-language reasoning for social robot navigation through efficient spatial reasoning learning, 2025.
\newblock URL \url{https://arxiv.org/abs/2503.07557}.

\bibitem[Li et~al.(2024)Li, Hogg, and Cohn]{stepgame-li-24}
Fangjun Li, David~C. Hogg, and Anthony~G. Cohn.
\newblock Advancing spatial reasoning in large language models: an in-depth evaluation and enhancement using the stepgame benchmark.
\newblock In \emph{Proceedings of the Thirty-Eighth AAAI Conference on Artificial Intelligence and Thirty-Sixth Conference on Innovative Applications of Artificial Intelligence and Fourteenth Symposium on Educational Advances in Artificial Intelligence}, AAAI'24/IAAI'24/EAAI'24. AAAI Press, 2024.
\newblock ISBN 978-1-57735-887-9.
\newblock \doi{10.1609/aaai.v38i17.29811}.
\newblock URL \url{https://doi.org/10.1609/aaai.v38i17.29811}.

\bibitem[Li et~al.(2025)Li, Wu, Du, Liu, Nghiem, and Shi]{li2025surveystateartlarge}
Zongxia Li, Xiyang Wu, Hongyang Du, Fuxiao Liu, Huy Nghiem, and Guangyao Shi.
\newblock A survey of state of the art large vision language models: Alignment, benchmark, evaluations and challenges, 2025.
\newblock URL \url{https://arxiv.org/abs/2501.02189}.

\bibitem[Minaee et~al.(2025)Minaee, Mikolov, Nikzad, Chenaghlu, Socher, Amatriain, and Gao]{minaee2025largelanguagemodelssurvey}
Shervin Minaee, Tomas Mikolov, Narjes Nikzad, Meysam Chenaghlu, Richard Socher, Xavier Amatriain, and Jianfeng Gao.
\newblock Large language models: A survey, 2025.
\newblock URL \url{https://arxiv.org/abs/2402.06196}.

\bibitem[Mirzaee \& Kordjamshidi(2022)Mirzaee and Kordjamshidi]{SpatialRole}
Roshanak Mirzaee and Parisa Kordjamshidi.
\newblock Transfer learning with synthetic corpora for spatial role labeling and reasoning.
\newblock 10 2022.
\newblock \doi{10.48550/arXiv.2210.16952}.

\bibitem[Noever \& Burdick(2021)Noever and Burdick]{puzzle-lang-noever-21}
David Noever and Ryerson Burdick.
\newblock Puzzle solving without search or human knowledge: An unnatural language approach.
\newblock \emph{CoRR}, abs/2109.02797, 2021.
\newblock URL \url{https://arxiv.org/abs/2109.02797}.

\bibitem[{Qwen Team}(2024)]{qwen2.5-1.5b-instruct}
{Qwen Team}.
\newblock Qwen2.5-1.5b-instruct.
\newblock \url{https://huggingface.co/Qwen/Qwen2.5-1.5B-Instruct}, 2024.
\newblock Hugging Face model card, accessed 2025-08-10.

\bibitem[Szot et~al.(2024)Szot, Mazoure, Agrawal, Hjelm, Kira, and Toshev]{szot2024groundingmultimodallargelanguage}
Andrew Szot, Bogdan Mazoure, Harsh Agrawal, Devon Hjelm, Zsolt Kira, and Alexander Toshev.
\newblock Grounding multimodal large language models in actions, 2024.
\newblock URL \url{https://arxiv.org/abs/2406.07904}.

\bibitem[Todd et~al.(2023)Todd, Earle, Nasir, Green, and Togelius]{Todd_2023}
Graham Todd, Sam Earle, Muhammad~Umair Nasir, Michael~Cerny Green, and Julian Togelius.
\newblock Level generation through large language models.
\newblock In \emph{Proceedings of the 18th International Conference on the Foundations of Digital Games}, FDG 2023, pp.\  1–8. ACM, April 2023.
\newblock \doi{10.1145/3582437.3587211}.
\newblock URL \url{http://dx.doi.org/10.1145/3582437.3587211}.

\bibitem[Vaswani et~al.(2017)Vaswani, Shazeer, Parmar, Uszkoreit, Jones, Gomez, Kaiser, and Polosukhin]{att_all_you_nead}
Ashish Vaswani, Noam Shazeer, Niki Parmar, Jakob Uszkoreit, Llion Jones, Aidan~N. Gomez, \L{}ukasz Kaiser, and Illia Polosukhin.
\newblock Attention is all you need.
\newblock In \emph{Proceedings of the 31st International Conference on Neural Information Processing Systems}, NIPS'17, pp.\  6000–6010, Red Hook, NY, USA, 2017. Curran Associates Inc.
\newblock ISBN 9781510860964.

\bibitem[Wang et~al.(2024)Wang, Luo, Wang, and Yan]{wang2024bothumandetectingchatgpt}
Hong Wang, Xuan Luo, Weizhi Wang, and Xifeng Yan.
\newblock Bot or human? detecting chatgpt imposters with a single question, 2024.
\newblock URL \url{https://arxiv.org/abs/2305.06424}.

\bibitem[Wang et~al.(2025)Wang, Hooi, Wang, Yang, Huang, and Cai]{wang2025textspeaksloudervision}
Zhaochen Wang, Bryan Hooi, Yiwei Wang, Ming-Hsuan Yang, Zi~Huang, and Yujun Cai.
\newblock Text speaks louder than vision: Ascii art reveals textual biases in vision-language models, 2025.
\newblock URL \url{https://arxiv.org/abs/2504.01589}.

\bibitem[Wu et~al.(2024{\natexlab{a}})Wu, Zhong, Xing, Lai, Liu, Chen, Wang, Zhu, Lu, Lu, Luo, Qiao, and Dai]{vis-lan-wu-24}
Jiannan Wu, Muyan Zhong, Sen Xing, Zeqiang Lai, Zhaoyang Liu, Zhe Chen, Wenhai Wang, Xizhou Zhu, Lewei Lu, Tong Lu, Ping Luo, Yu~Qiao, and Jifeng Dai.
\newblock Visionllm v2: an end-to-end generalist multimodal large language model for hundreds of vision-language tasks.
\newblock In \emph{Proceedings of the 38th International Conference on Neural Information Processing Systems}, NIPS '24, Red Hook, NY, USA, 2024{\natexlab{a}}. Curran Associates Inc.
\newblock ISBN 9798331314385.

\bibitem[Wu et~al.(2024{\natexlab{b}})Wu, Mao, Zhang, Xia, Dong, Cui, and Wei]{wu2024mindseyellmsvisualizationofthought}
Wenshan Wu, Shaoguang Mao, Yadong Zhang, Yan Xia, Li~Dong, Lei Cui, and Furu Wei.
\newblock Mind's eye of llms: Visualization-of-thought elicits spatial reasoning in large language models, 2024{\natexlab{b}}.
\newblock URL \url{https://arxiv.org/abs/2404.03622}.

\bibitem[Yao et~al.(2023)]{yao2023tot}
Shunyu Yao et~al.
\newblock Tree of thoughts: Deliberate problem solving with large language models.
\newblock \emph{arXiv preprint arXiv:2305.10601}, 2023.

\bibitem[Zhang et~al.(2024)Zhang, Ma, Li, Qiao, Wang, Chai, Wu, Bansal, and Kordjamshidi]{zhang2024visionandlanguagenavigationtodaytomorrow}
Yue Zhang, Ziqiao Ma, Jialu Li, Yanyuan Qiao, Zun Wang, Joyce Chai, Qi~Wu, Mohit Bansal, and Parisa Kordjamshidi.
\newblock Vision-and-language navigation today and tomorrow: A survey in the era of foundation models, 2024.
\newblock URL \url{https://arxiv.org/abs/2407.07035}.

\end{thebibliography}
\bibliographystyle{colm2025_conference}

\appendix
\section{Environment Setup Example}
\label{appendix-A}

\label{app:responses}

\begin{tcolorbox}[
    colback=gray!10,
    colframe=gray!50,
    colbacktitle=blue!15,
    coltitle=blue, 
    arc=3mm,
    boxrule=0.5pt,
    width=1\textwidth,
    center, 
    title = System prompt for Dynamic Setting,
    label=box:dynamic   
]

\begin{lstlisting}[basicstyle=\ttfamily\footnotesize]
You are an expert at analyzing ASCII art shapes. Two shapes are provided: 
Shape A (the target) and Shape B (the current state), separated by '%$%$%$%'.
Each line of the shapes starts and ends with a star (*) character.

Your goal is to transform Shape B into Shape A 
by analyzing one type of transformation at a time.

You may analyze rotation, translation, or scaling --- but try to analyze 
DIFFERENT types rather than repeating.

You have already analyzed: {analyzed}

IMPORTANT: Choose a transformation type you haven't analyzed yet if possible.

TASK 1 - ROTATION:
If analyzing rotation: Determine what rotation is needed to transform Shape B 
into Shape A.
Classify using exactly one of these labels:
  - 'no_rotation': The shapes have the same orientation (0 rotation needed)
  - 'quarter_rotation': Shape B needs a 90 degrees rotation to match Shape A
  - 'slight_rotation': Shape B needs a small rotation (<90 degrees) 
  to match Shape A

TASK 2 - TRANSLATION:
If analyzing translation: Determine how Shape B must be moved to match Shape
A's position.
Classify using exactly one of these labels:
  - 'no_translation': Shapes are already in the same position
  - 'up': Shape B must be moved up to match Shape A
  - 'down': Shape B must be moved down to match Shape A
  - 'left': Shape B must be moved left to match Shape A
  - 'right': Shape B must be moved right to match Shape A

TASK 3 - SCALING:
If analyzing scaling: Determine the size adjustment needed to transform Shape B
to match Shape A.
Classify using exactly one of these labels:
  - 'no_scaling': Both shapes have the same size
  - 'double_size': Shape B is half size and must be enlarged to match Shape A
  - 'half_size': Shape B is twice as large and must be shrunk to match Shape A

INSTRUCTIONS:
1. Identify which transformation type would be most useful to analyze now
2. Carefully compare both ASCII art shapes
3. Determine the transformation needed to convert Shape B into Shape A
4. Respond with the appropriate label inside <answer></answer> tags 

\end{lstlisting}

\end{tcolorbox}
\newpage

\begin{tcolorbox}[
    colback=gray!10,
    colframe=gray!50,
    colbacktitle=blue!15,
    coltitle=blue, 
    arc=3mm,
    boxrule=0.5pt,
    width=1\textwidth,
    center, 
    title = System prompt for Static Setting,
    label=box:static 
]

\begin{lstlisting}[basicstyle=\ttfamily\footnotesize]
You are an expert at analyzing ASCII art shapes. Two shapes are provided: 
Shape A (the target) and Shape B (the current state), separated by '%$%$%$%'.
Each line of the shapes starts and ends with a star (*) character.

Your goal is to transform Shape B into Shape A 
by analyzing one type of transformation at a time.

You may analyze rotation, translation, or scaling --- but try to analyze 
DIFFERENT types rather than repeating.

CRITICAL INSTRUCTION: THE MAP IS NOT UPDATING 
The Shape B shown below is the **STATIC INITIAL STATE**
It does **NOT** reflect the moves you have already made.
You must RELY ON YOUR MEMORY of the following actions you have already performed:
HISTORY OF ACTIONS: [{analyzed}]

To solve this:
1. Look at the Initial Shape B.
2. Mentally apply the 'HISTORY OF ACTIONS' to it to imagine the *current* state.
3. Determine the NEXT step needed from that imagined state.
4. Choose a transformation type you haven't analyzed yet if possible.

TASK 1 - ROTATION:
If analyzing rotation: Determine what rotation is needed to transform Shape B 
into Shape A.
Classify using exactly one of these labels:
  - 'no_rotation': The shapes have the same orientation (0 rotation needed)
  - 'quarter_rotation': Shape B needs a 90 degrees rotation to match Shape A
  - 'slight_rotation': Shape B needs a small rotation (<90 degrees) 
  to match Shape A

TASK 2 - TRANSLATION:
If analyzing translation: Determine how Shape B must be moved to match Shape
A's position.
Classify using exactly one of these labels:
  - 'no_translation': Shapes are already in the same position
  - 'up': Shape B must be moved up to match Shape A
  - 'down': Shape B must be moved down to match Shape A
  - 'left': Shape B must be moved left to match Shape A
  - 'right': Shape B must be moved right to match Shape A

TASK 3 - SCALING:
If analyzing scaling: Determine the size adjustment needed to transform Shape B
to match Shape A.
Classify using exactly one of these labels:
  - 'no_scaling': Both shapes have the same size
  - 'double_size': Shape B is half size and must be enlarged to match Shape A
  - 'half_size': Shape B is twice as large and must be shrunk to match Shape A

INSTRUCTIONS:
1. Identify which transformation type would be most useful to analyze now
2. Carefully compare both ASCII art shapes
3. Determine the transformation needed to convert Shape B into Shape A
4. Respond with the appropriate label inside <answer></answer> tags
\end{lstlisting}

\end{tcolorbox}
\newpage

\begin{tcolorbox}[
    colback=gray!10,
    colframe=gray!50,
    colbacktitle=blue!15,
    coltitle=blue, 
    arc=3mm,
    boxrule=0.5pt,
    width=1\textwidth,
    center,
     title=Sample Environment,
     label=box:env   
]
\label{app:sample-env}
\begin{lstlisting}
TARGET (Shape A):
*                           *
*                           *
*                           *
*                           *
*                           *
*                           *
*                           *
*                           *
*                           *
*                           *
*                           *
*                           *
*                           *
*                           *
*                           *
*                           *
*                           *
*                           *
*                           *
*                        ## *
*                      ##  #*
*                    ##     *
*                  ##     ##*
*                ##     ##  *
*                 ##  ##    *
*                   ##      *
*                           *
%$%$%$%
CURRENT (Shape B):
*                           *
*                           *
*            #              *
*          ## ##            *
*           #   #           *
*            ## ##          *
*              #            *
*                           *
*                           *
*                           *
*                           *
*                           *
*                           *
*                           *
*                           *
*                           *
*                           *
*                           *
*                           *
*                           *
*                           *
*                           *
*                           *
*                           *
*                           *
*                           *
*                           *
------------------------------
\end{lstlisting}

\end{tcolorbox}

\newpage

\begin{tcolorbox}[
    colback=gray!10,
    colframe=gray!50,
    colbacktitle=blue!15,
    coltitle=blue, 
    arc=3mm,
    boxrule=0.5pt,
    width=1.1\textwidth,
    center,
    title= Model Responses (Dynamic setting with State Updates),
    label=box:response   
]

\textbf{$\star$ Ground-Truth:} double size, quarter rotation, right, down, down.

\vspace{0.5em}
\noindent
\textbf{$\times$ Qwen-Instruct:} slight rotation, slight rotation, slight rotation, slight rotation, slight rotation.

\vspace{0.5em}
\noindent
\textbf{$\times$ Qwen-Physics:} slight rotation, quarter rotation, slight rotation, slight rotation, slight rotation.

\vspace{0.5em}
\noindent
\textbf{$\times$ Qwen-DirectRL:} double size, quarter rotation, down, down, up.

\vspace{0.5em}
\noindent
\textbf{$\checkmark$ Qwen-PhysRL (ours):} double size, quarter rotation, right, down, down.

\end{tcolorbox}

\section{Random Policy Reward Calculation}
\label{app:rand}
To establish a baseline for the random policy $\pi_{\text{rnd}}$, we analytically derive the expected reward $E[R_k]$ at step $k$. An action $a$ yields a positive reward only if its usage count $N_a$ over the preceding $k\!-\!1$ steps does not exceed its ground-truth quota $C_a$. We model $N_a$ as a binomial random variable, $N_a \sim \mathrm{Binomial}(k-1, 1/|\mathcal{A}|)$, reflecting uniform random action selection. The expected reward at step $k$ is therefore given by
\begin{equation}
E[R_k] = \sum_{a \in GT} r(a)\,\mathbb{P}(N_a < C_a)
\;-\;
\lambda \!\left( 1 - \sum_{a \in GT} \mathbb{P}(N_a < C_a) \right),
\end{equation}
where $GT$ denotes the set of ground-truth required actions, $r(a)$ is the reward associated with a valid action, and $\lambda$ is the penalty incurred for an invalid action. The probability term expands as the cumulative binomial sum
\begin{equation}
\mathbb{P}(N_a < C_a) = \sum_{i=0}^{C_a-1} \binom{k-1}{i}
\left(\frac{1}{|\mathcal{A}|}\right)^i
\left(1-\frac{1}{|\mathcal{A}|}\right)^{k-1-i}.
\end{equation}
For the subset of samples considered in our evaluation, the ground-truth action quotas consist of two occurrences of one translation direction ($C=2$), one occurrence of another translation direction ($C=1$), and a single occurrence each for rotation and scaling ($C=1$). The expected cumulative reward up to step $K$ then follows directly as $\sum_{j=1}^{K} E[R_j]$.

\end{document}